\documentclass[10pt,twocolumn,letterpaper]{article}

\usepackage{cvpr}         

\usepackage[accsupp]{axessibility}  %

\usepackage{graphicx}
\usepackage{algorithm}
\usepackage{algorithmic}
\usepackage{diagbox}
\usepackage{multirow}
\usepackage{subcaption}
\usepackage{booktabs}
\usepackage{siunitx}
\usepackage{threeparttable}
\usepackage{wrapfig}
\usepackage{caption}


\definecolor{cvprblue}{rgb}{0.21,0.49,0.74}
\usepackage[pagebackref,breaklinks,colorlinks,allcolors=cvprblue]{hyperref}

\def\paperID{12075} 
\def\confName{CVPR}
\def\confYear{2026}

\title{Physical Adversarial Clothing Evades Visible-Thermal Detectors via Non-Overlapping RGB-T Pattern}

\author{
  Xiaopei Zhu\textsuperscript{1 *} \quad
  Guanning Zeng\textsuperscript{1 *} \quad
  Zhanhao Hu\textsuperscript{2} \quad
  Jun Zhu\textsuperscript{1,3 †} \quad
  Xiaolin Hu\textsuperscript{1,3,4 †} \\
  \textsuperscript{1}Department of Computer Science and Technology, BNRist, Tsinghua University \\
  \textsuperscript{2}University of California Berkeley \\
  \textsuperscript{3}IDG/McGovern Institute for Brain Research, Tsinghua University \\
  \textsuperscript{4}Chinese Institute for Brain Research (CIBR)
}
\begin{document}
\maketitle
\begingroup
\renewcommand\thefootnote{\fnsymbol{footnote}}
\footnotetext[1]{\footnotesize Equal contribution.}
\footnotetext[2]{\footnotesize Corresponding authors.}
\endgroup

\begin{abstract}
Visible-thermal (RGB-T) object detection is a crucial technology for applications such as autonomous driving, where multimodal fusion enhances performance in challenging conditions like low light. However, the security of RGB-T detectors, particularly in the physical world, has been largely overlooked. This paper proposes a novel approach to RGB-T physical attacks using adversarial clothing with a non-overlapping RGB-T pattern (NORP). To simulate full-view (0$^{\circ}$–360$^{\circ}$) RGB-T attacks, we construct 3D RGB-T models for human and adversarial clothing. NORP is a new adversarial pattern design using distinct visible and thermal materials without overlap, avoiding the light reduction in overlapping RGB-T patterns (ORP). To optimize the NORP on adversarial clothing, we propose a spatial discrete-continuous optimization (SDCO) method. We systematically evaluated our method on RGB-T detectors with different fusion architectures, demonstrating high attack success rates both in the digital and physical worlds. Additionally, we introduce a fusion-stage ensemble method that enhances the transferability of adversarial attacks across unseen RGB-T detectors with different fusion architectures.
Our code is available at: \url{https://github.com/zxp555/RGBT-Clothing}.
\end{abstract}

\section{Introduction}

Visible-thermal (RGB-T) object detection is a type of multimodal object detection with important applications in fields such as autonomous driving and medical AI \citep{RGBT_ADAS,lai2024regseg,dasgupta2022spatio}. For example, in challenging conditions such as adverse weather or nighttime, RGB-T object detection leverages thermal imaging to compensate for the performance degradation of visible-only detectors while producing clearer predictions than thermal-only detectors by preserving more details from visible-light images. This advantage significantly enhances the robustness of autonomous driving systems across diverse scenarios. Based on different fusion strategies for multimodal information, RGB-T object detectors can be classified into four categories: early-fusion, mid-fusion, late-fusion, and independent visible and thermal detectors.

\begin{figure}[tbp]
\centering
\includegraphics[width=1\columnwidth]{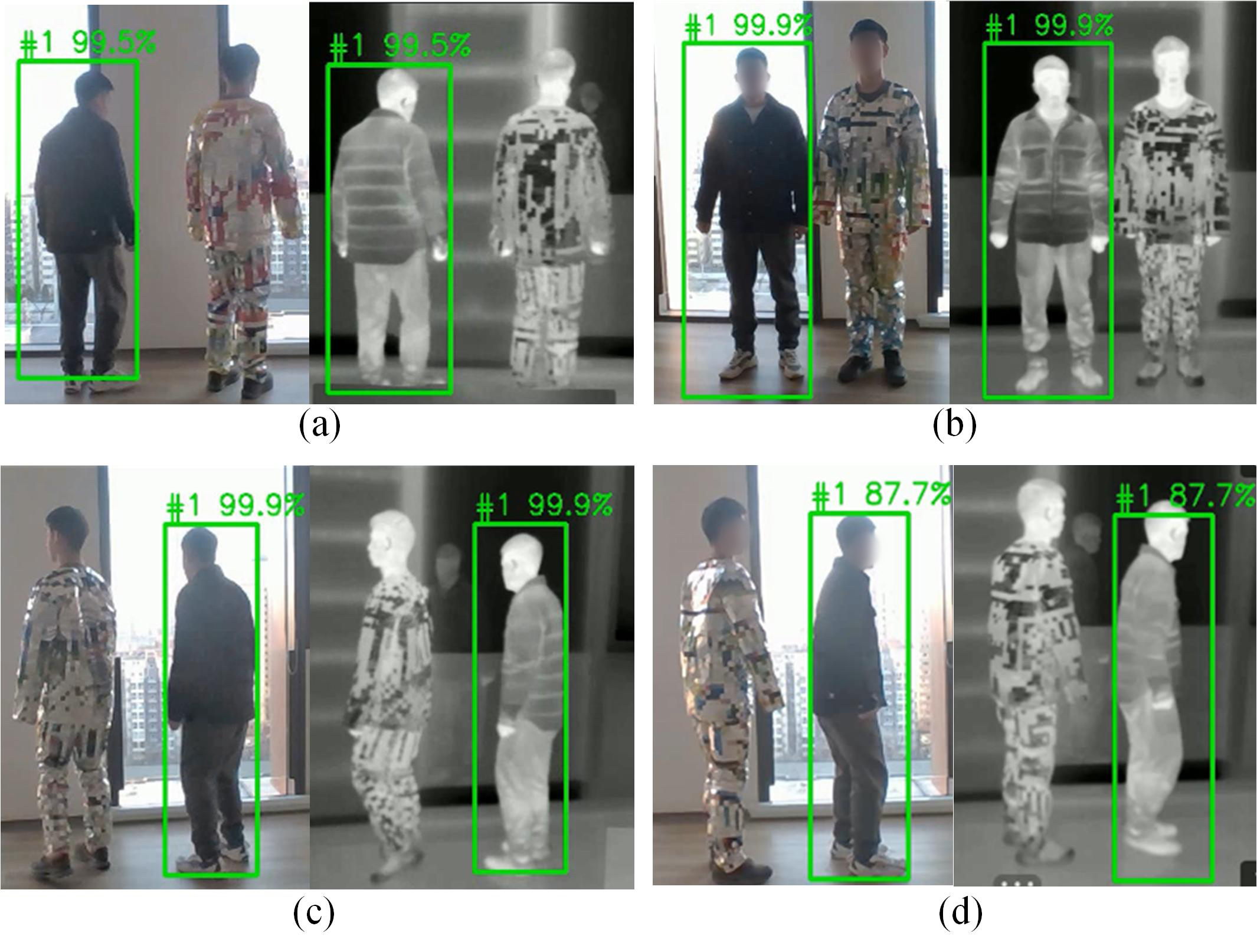} 
\caption{ Demonstration of physical attacks against RGB-T detectors with different fusion architectures: (a) early-fusion, (b) mid-fusion, (c) late-fusion, and (d) independent RGB-T detectors at multiple views. The individual wearing adversarial clothes remained undetected in both modalities, while the individual wearing ordinary clothes was detected (indicated by a bounding box). See \textit{Supplementary Material (SM)} for the Demo Video. }
\label{one_example}
\end{figure} 

Despite the widespread applications of RGB-T detectors, their security has received little attention because multimodal detectors are commonly assumed to be relatively robust. However, this is very important for the safety of AI systems in real-world applications, such as autonomous driving. To address this issue, physical adversarial examples \citep{wei2024physical} offer an effective approach to identifying vulnerabilities in AI systems deployed in the physical world and inspiring novel defense strategies. Currently, most physical adversarial examples focuses either solely on the visible modality \citep{conf/cvpr/ThysRG19,Xu2020ECCVshirt,wu2020making,hu2021naturalistic,Hu_2022_CVPR,Hu_2023_CVPR,wei2022adversarial} or the thermal modality \citep{ZHU_2021_AAAI,ZHU_2022_CVPR,ZHU_2024_CVPR,wei2023hotcold,wei2023physically}. Due to the significant differences in imaging mechanisms between these two modalities, adversarial examples crafted for one modality cannot be effectively transferred to the other. 
As a result, these physical adversarial examples lack the ability to attack multimodal detectors.

The earliest physical adversarial example in visible-thermal settings, AdvB~\citep{ZHU_2021_AAAI,zhu2024hiding}, combined RGB and thermal adversarial strategies. More recently, MAP~\citep{kim2022map}, UAP~\citep{wei2023unified}, and MIC~\citep{kim2023multispectral} have further advanced physical adversarial attacks in this domain. 
However, they have two major limitations. First, AdvB,
MAP, and UAP are realized as 2D patches, which can only attack detectors at a narrow range of viewing angles (e.g., -30$^{\circ}$ to 30$^{\circ}$). Second, MIC deploys an overlapping RGB-T pattern (ORP) by attaching multiple special low-E films onto a printed fabric, which diminishes the visibility of the printed adversarial pattern and increases production costs. These limitations result in the vulnerability of RGB-T detectors across different physical settings not being fully explored.

To address the first limitation, 
we construct aligned 3D RGB-T models of the human body and clothing to simulate full-view (0$^{\circ}$–360$^{\circ}$) attacks in the digital world, and manufacture corresponding 3D RGB-T adversarial clothing to enable full-view attacks in the physical world.
To address the second limitation, 
we propose a non-overlapping RGB-T pattern (NORP) design for RGB-T adversarial clothing.
NORP not only leverages the visible and thermal properties of different materials simultaneously but also ensures that these materials do not overlap, thereby avoiding the light reduction problem inherent to ORP. Moreover, we use commonly available materials (fabric and aluminum film) to deploy NORP, which offers the advantage of low cost.

A key challenge of optimizing NORP lies in the spatial dependency between the optimization variables of the visible and thermal patterns. For example, when a pixel is chosen to be aluminum film, its RGB values are fixed and cannot be simultaneously optimized as continuous variables, and vice versa. To address this challenge, we propose a spatial discrete-continuous optimization (SDCO) method that enables simultaneous optimization of continuous RGB pixels and discrete thermal pixels within NORP.

Different from previous methods \citep{kim2022map,wei2023unified,kim2023multispectral} that validate on only one type of (e.g., mid-fusion) RGB-T detector, we systematically evaluate our approach on RGB-T detectors with various fusion architectures, both in the digital and physical world.
Experiments show that our RGB-T adversarial clothes effectively attacked RGB-T detectors across multiple fusion architectures, with average ASR of 99.6\% and 71.0\% in the digital and physical worlds, respectively. Fig. \ref{one_example} shows several visualized examples. Furthermore, we propose a fusion-stage ensemble method that enables a piece of adversarial clothing to simultaneously attack RGB-T detectors with multiple fusion architectures. 
Our work comprehensively reveals the vulnerabilities of RGB-T detectors across different fusion architectures, viewing angles, and distances, which is important for building more robust detectors in the future.

\section{Related Work}
\subsection{RGB-T Object Detection}

RGB-T object detectors are multimodal object detectors that integrate visible-light and thermal imaging modalities.
Those detectors are classified into early-fusion detectors \cite{liu2016multispectral, chen2022multimodal, el2023enhanced, zhang2024e2e, zhao2023metafusion, liu2022target} which integrate multimodal information at the image level, mid-fusion detectors \cite{Zhang_2019_ICCV, chen2022multimodal, hu2022dmffnet, xiang2022rgb, DBLP:conf/eccv/GuoGLMG24, dong2024fusion, shen2024icafusion, xing2024ms} which focus on feature-level fusion, late-fusion detectors \cite{liu2016multispectral, ni2022modality, chen2022multimodal, wagner2016multispectral} which merge multimodal information at the prediction stage, and independent visible and thermal detectors \cite{khanam2024yolov11, wang2023yolov7, wang2024yolov9, zhu2020deformable}, where each modality operates independently. 

\subsection{Physical Attacks in Visible Modality}

Most physical attacks focus on the visible modality, given the prevalence of RGB cameras. Researchers typically design pixel-level adversarial patterns and then print them onto papers \cite{conf/cvpr/ThysRG19,wu2020making}, stickers \cite{wei2022adversarial,DBLP:conf/cvpr/DuanM00QY20}, or clothing \cite{Xu2020ECCVshirt,hu2021naturalistic,Hu_2022_CVPR,Hu_2023_CVPR}. For instance, 
Duan et al. \cite{DBLP:conf/cvpr/DuanM00QY20} proposed an adversarial sticker to hide from visible detectors. Hu et al. \cite{Hu_2023_CVPR} presented physical adversarial clothing to evade person detectors. However, due to the differences in imaging mechanisms, adversarial patterns designed for the visible modality cannot be effectively displayed in the thermal modality.

\subsection{Physical Attacks in Thermal Modality}

A few works focus on thermal physical attacks. Researchers typically begin by mathematically modeling the thermal characteristics of specific devices or materials, then optimize key parameters and deploy the solution physically. Physical deployment methods include both heating devices \cite{ZHU_2021_AAAI, wei2023hotcold} and thermal insulation materials \cite{ZHU_2022_CVPR,wei2023physically,ZHU_2024_CVPR}. For example, Zhu et al. \cite{ZHU_2021_AAAI} proposed an adversarial bulb board that evades multiple thermal detectors. 
Wei et al. \cite{wei2023physically} introduced aerogel-based adversarial patches to attack thermal detectors. Similarly, thermal adversarial patterns cannot be effectively displayed in the visible modality.

\begin{figure*}[htbp]
\centering
\includegraphics[scale=0.69]{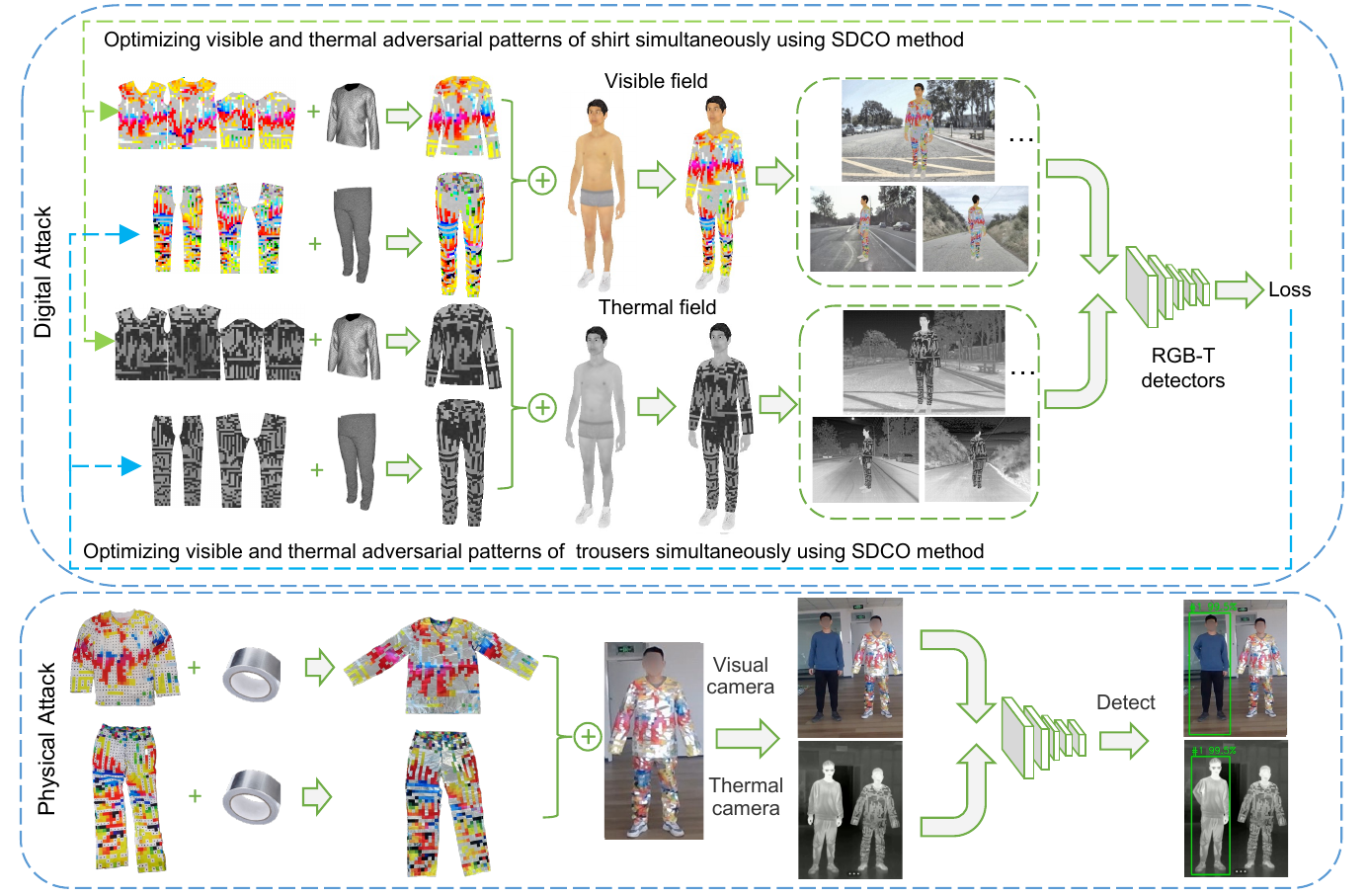} 
\caption{The overall pipeline of the proposed method. We jointly optimizes visible and thermal patterns on 3D RGB-T clothing models by spatial discrete-continuous optimization method, whose attack performance are then rendered and evaluated by RGB-T detectors.}
\label{main_process}
\end{figure*} 

\subsection{Physical Attacks in Visible-Thermal Modality}

Only a limited number of works focus on visible-thermal physical attacks. 
Such works require not only the simultaneous optimization of visible and thermal variables but also the physical implementation of both modalities. 
Zhu et al. \cite{ZHU_2021_AAAI,zhu2024hiding} proposed the first RGB-T physical adversarial attack method AdvB based on small bulbs and printed paper.
Kim et al. \cite{kim2022map} proposed a multispectral adversarial patch (MAP) to attack a mid-fusion RGB-T person detector. Later, they \cite{kim2023multispectral} introduced a low-E film–based clothing, named MIC, to attack the same RGB-T detector. Wei et al. \cite{wei2023unified} developed a unified adversarial patch (UAP) to evade independent RGB-T detectors. 
However, these methods face certain limitations. For example, AdvB, MAP and UAP exhibit restricted attack angles (e.g., -30$^{\circ}$ to 30$^{\circ}$), while MIC deploys an overlapping RGB-T pattern with low-E films that reduce light transmittance by 30\%, thereby diminishing the visibility of the printed fabric pattern. These limitations result in the vulnerability of RGB-T detectors across different physical settings not being fully explored.

\section{Method}

\subsection{Problem Formulation}

For a given target $x$, its visible image is denoted as $x_{\mathrm{vis}}$, while its thermal image is denoted as $x_{\mathrm{thm}}$. After adding adversarial perturbations, the adversarial images in both modalities are represented as $x_{\mathrm{vis}}^{\mathrm{adv}}$  and $x_{\mathrm{thm}}^{\mathrm{adv}}$, respectively. The objective of an RGB-T attack is to ensure that the RGB-T detector $f$ fails to detect the target $x$. Given a detection threshold $q$, the optimization objective is:
\begin{equation}
\label{equ:detector}
f(x_{\mathrm{vis}}^{\mathrm{adv}},x_{\mathrm{thm}}^{\mathrm{adv}})<q.
\end{equation}

Our RGB-T attack pipeline is shown in Fig. \ref{main_process}. First, we construct aligned 3D RGB-T models for both the human body and clothing. Then we design the non-overlapping RGB-T pattern (NORP) and apply our SDCO method to optimize the visible and thermal adversarial patterns simultaneously. Finally, based on the optimization results, we manufacture the multimodal physical clothes and evaluate their performance in the physical world.

\subsection{Building 3D RGB-T Models}

\label{sec:building_3D}
To simulate full-view RGB-T attacks, we need to construct an aligned 3D RGB-T model. Since most existing 3D models are RGB models, we develop a method for extending an 3D RGB model into an aligned 3D RGB-T model based on a previous thermal 3D modeling approach \cite{ZHU_2024_CVPR}. Initially, we utilized publicly available 3D RGB human and clothing models \cite{Hu_2023_CVPR} as the foundation. Next, we take a clothing model as an example to illustrate our method.

The challenge in building an aligned 3D RGB-T model lies in generating an thermal ``skin" that aligns with the 3D mesh model. To address this, we first unfold the faces of the 3D mesh model into a 2D faces map and organize it into different regions, such as the back and arms, using Maya software. Next, we capture real thermal images of clothing using an thermal camera and process them to align with the faces map, producing an aligned thermal texture map. See \textit{Supplementary Material (SM)} for how these photos were captured and processed. This process ensures that the real thermal texture is properly aligned with the 3D mesh model, and the final rendered 3D RGB-T models are shown in \textit{SM}.

\subsection{Designing Non-Overlapping RGB-T Pattern}

\label{sec:design}
After constructing the 3D model of the clothing, we propose to design non-overlapping RGB-T pattern (NORP) for the multimodal adversarial clothing. 
Since we use printable fabric and aluminum film to deploy NORP onto adversarial clothing, each location on the clothing is either printed with RGB colors or covered with aluminum film, but not both.
We parameterize the NORP, so that it can be represented by $N$ pixels.
Let \( X =[X_i]= [r_i, g_i, b_i, t_i]_{i = 1, 2, \dots, N} \) represent the NORP, where \( [r_i, g_i, b_i] \) denotes the RGB value of visible light and \( t_i \) denotes the thermal emission intensity.

If the clothing at the pixel $i$ is covered with an aluminum film, its RGB-T value is fixed by $[r^{(\mathrm{T})}, g^{(\mathrm{T})}, b^{(\mathrm{T})}, t^{(\mathrm{film})}]$, which is determined by the characteristic of the aluminum film. Otherwise, the fabric can be printed with any RGB color, and the body temperature determines the thermal intensity. Therefore, its RGB-T value can be represented by $[r_i^{(\mathrm{V})}, g_i^{(\mathrm{V})}, b_i^{(\mathrm{V})}, t_i^{(\mathrm{body})}]$, 
where $t_i^{(\mathrm{body})}$ is sampled from the thermal texture.
Furthermore, we use an additional variable $p_i=0$ to represent that a pixel $i$ is covered with an aluminum film, and $p_i=1$ otherwise. The RGB-T pixel $X_i=[r_i, g_i, b_i, t_i]$ thus can be represented by
\begin{equation} 
\label{equ:constraints}
\begin{split}
    X_i = \ \mathrm{H}(Y_i)
    = \ p_i \cdot 
    [r^\mathrm{(V)}_i, g^\mathrm{(V)}_i, b^\mathrm{(V)}_i, t_i^\mathrm{(body)}] + \\
    (1-p_i) \cdot 
    [r^\mathrm{(T)}, g^\mathrm{(T)}, b^\mathrm{(T)}, t^\mathrm{(film)}],
\end{split}
\end{equation}
where $Y_i=[r^\mathrm{(V)}_i, g^\mathrm{(V)}_i, b^\mathrm{(V)}_i, p_i]$ is a collection of learnable variables.
Please note that the values of $r^\mathrm{(T)}$, $g^\mathrm{(T)}$, $b^\mathrm{(T)}$, $t_i^{(\mathrm{body})}$, $t^\mathrm{(film)}$ are measured values (constants).

\begin{figure}[tbp]
\centering
\includegraphics[width=1\columnwidth]{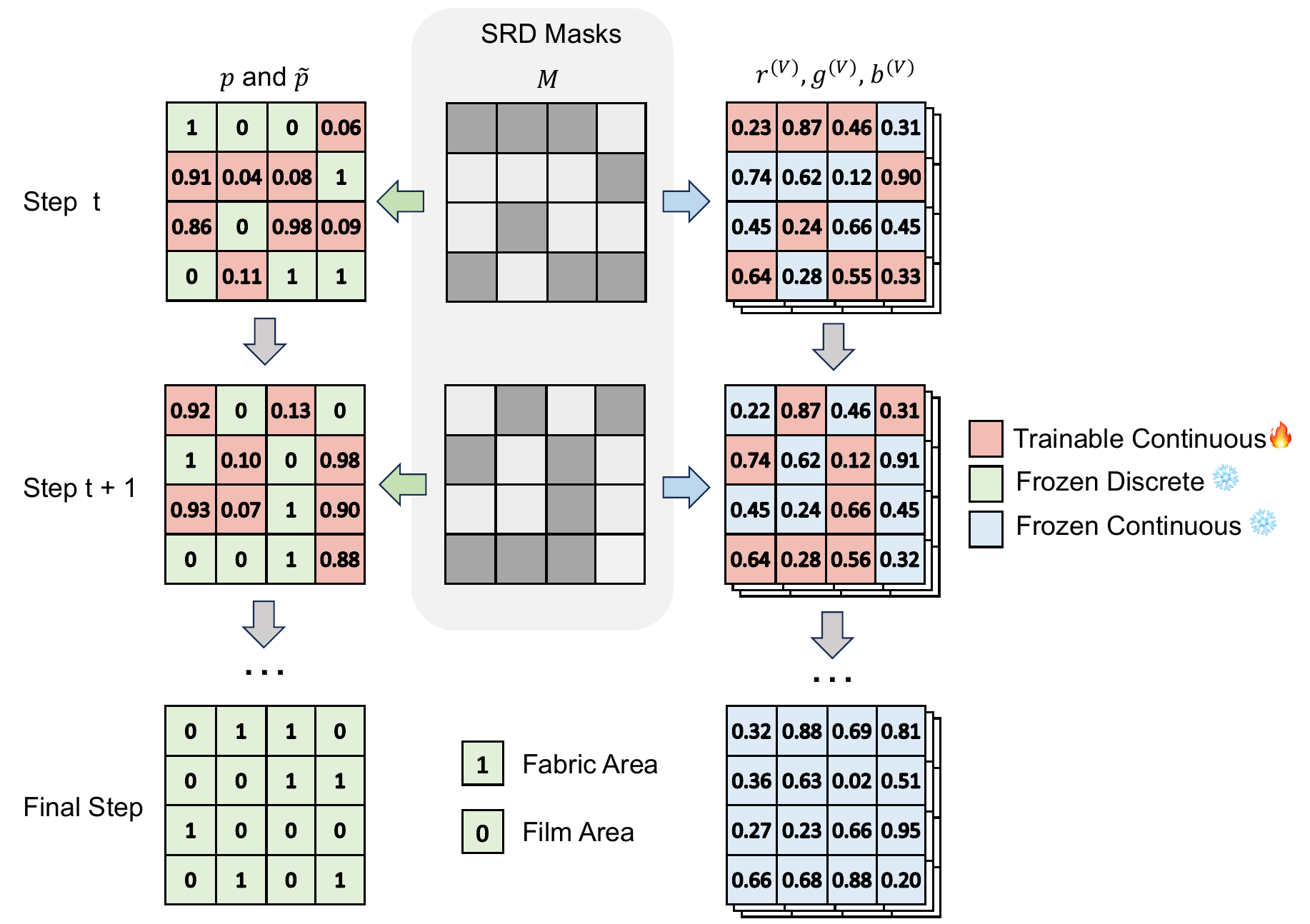} 
\caption{Illustration of the SDCO method. In SRD, black pixels represent discretizing $\tilde{p}$ and updating $r^\mathrm{(V)}, g^\mathrm{(V)}, b^\mathrm{(V)}$, while gray pixels represent freezing $r^\mathrm{(V)}, g^\mathrm{(V)}, b^\mathrm{(V)}$ and updating $\tilde{p}$. At final step, all $\tilde{p}$ are discretized, and NORP optimization is finished.}
\label{DCMA}
\end{figure} 

\subsection{Optimizing the Adversarial RGB-T Pattern} \label{sec:DCMA}

We develop a gradient-based algorithm to optimize the variables of NORP. Since $p_i$ is discrete in Equation \ref{equ:constraints}, we optimize a continuous variable $\tilde{p_i}$ instead, and discretize $\tilde{p_i}$ as $p_i= \mathbf{1}(\tilde{p}_i \geq 0.5)$ after the optimization. However, a naive method to optimize the variables $[r^\mathrm{(V)}_i, g^\mathrm{(V)}_i, b^\mathrm{(V)}_i, \tilde{p}_i]$ directly led to suboptimal results (see Sec. \ref{sec:ablation} for details). This is because the continuous variables $r^\mathrm{(V)}_i$, $g^\mathrm{(V)}_i$, $b^\mathrm{(V)}_i$ and the discrete variable $p_i$ are entangled, thus the approximation of $p_i$ affects other variables as well.

Therefore, we propose a spatial discrete-continuous optimization (SDCO) method based on Spatially-Random Discretization (SRD). The core idea of SRD is to discretize some pixels during the gradient optimization process, while computing gradients for other pixels that remain continuous. As shown in Fig. \ref{DCMA}, during each iteration, we apply a random mask to discretize a portion of the thermal adversarial pattern's pixels and freeze these variables. The random discretization ratio is denoted as $\alpha$, with the corresponding visible variables in these regions being optimized. Conversely, the remaining $1-\alpha$ portion of the thermal adversarial pattern's pixels is set to be trainable, while the corresponding visible pixels are frozen. As the mask changes randomly throughout the iterations, each pixel has an equal probability of being in the trainable state, allowing for iterative updates.

This method balances the optimization of visible and thermal adversarial patterns while satisfying the spatial interdependency constraints between these two patterns specified in Equation \ref{equ:constraints}, ensuring that the multimodal adversarial pattern is physically realizable. The process of the algorithm is outlined in Algorithm \ref{alg:DCMA}.

\begin{algorithm}
\caption{Spatial Discrete-Continuous Optimization}
\label{alg:DCMA}
\begin{algorithmic}[1]
\REQUIRE Variables $Y = \{{[r^\mathrm{(V)}_i, g^\mathrm{(V)}_i, b^\mathrm{(V)}_i, p_i]}\}_{i=1}^N$, Continuous $\tilde{p}_i$, step size $\eta$, discretization probability $\alpha$, max iterations $T$, detection process $f(X)$ in Equation \ref{equ:detector}, function $\mathrm{H}(Y)$ in Equation \ref{equ:constraints}, loss function $\mathcal{L}$ in Equation \ref{equ:loss}
\STATE Initialize iteration $j \gets 0$
\WHILE{$j < T$}
    \STATE Generate random mask $M \in \{0,1\}^N$ where $M_i \sim \text{Bernoulli}(\alpha)$ \hfill \textit{\% Spatially-Random Discretization}
    \FOR{each pixel $i$}
        \IF{$M_i = 1$}  
            \STATE $p_i \gets \mathbf{1}(\tilde{p_i} \geq 0.5)$ \hfill \textit{\% Discretize thermal}
        \ELSE  
            \STATE $p_i \gets \tilde{p_i}$ \hfill \textit{\% Keep thermal continuous}
        \ENDIF
    \ENDFOR 
    \hfill \\ \textit{\% One single forward \& backward step}
    \STATE Compute adversarial loss $L \gets \mathcal{L}(f(\mathrm{H}(Y)))$ 
    \STATE Calculate gradients $\nabla_Y L \gets \frac{\partial L}{\partial Y}$ 
    \FOR{each pixel $i$}  
        \IF{$M_i = 1$}
            \STATE $\nabla_{p_i} L \gets 0$ \hfill \textit{\% Block thermal gradient }
        \ELSE
            \STATE $\nabla_{[r^\mathrm{(V)}_i, g^\mathrm{(V)}_i, b^\mathrm{(V)}_i]} L \gets 0$ \hfill \textit{\% Block RGB gradient}
        \ENDIF
    \ENDFOR
    \STATE Update parameters: $Y \gets Y - \eta \nabla_Y L$ 
    \STATE $j \gets j + 1$
\ENDWHILE
\STATE \textbf{Final Binarization:} $p_i \gets \mathbf{1}(\tilde{p_i} \geq 0.5)$ for all $i$ 
\RETURN Adversarial Texture $X = \mathrm{H}(Y)$
\end{algorithmic}
\end{algorithm}

\subsection{Applying the Adversarial RGB-T Patterns to 3D RGB-T Models} \label{sec:apply pattern}

Let $T_{\text{adv}}^{\text{vis}}$ and $T_{\text{adv}}^{\text{thm}}$ denote the optimized adversarial textures for the visible and thermal modalities, respectively. 
We first apply EOT algorithm \citep{conf/icml/AthalyeEIK18} to simulate physical perturbations during optimization.
The adversarial textures after the EOT transformation are represented as:

\begin{equation}
\begin{aligned}
T_{\mathrm{adv\text{-}E}}^{\mathrm{vis}} &= \mathrm{EOT}\!\left(T_{\mathrm{adv}}^{\mathrm{vis}}\right) \\
T_{\mathrm{adv\text{-}E}}^{\mathrm{thm}} &= \mathrm{EOT}\!\left(T_{\mathrm{adv}}^{\mathrm{thm}}\right)
\end{aligned}
\end{equation}

We then use the renderer $\mathrm{R}$ to map the adversarial textures onto the surface of the 3D clothing mesh $M_{\text{cloth}}$. The parameters $\phi$ of the renderer include rendering distances, angles, etc. The rendered adversarial clothing images in both visible and thermal modalities are represented as:

\begin{equation}
\begin{aligned}
I_{\mathrm{cloth}}^{\mathrm{vis}}&=\mathrm{R}(M_{\mathrm{cloth}},T_{\mathrm{adv\text{-}E}}^{\mathrm{vis}},\phi) \\ I_{\mathrm{cloth}}^{\mathrm{thm}}&=\mathrm{R}(M_{\mathrm{cloth}},T_{\mathrm{adv\text{-}E}}^{\mathrm{thm}},\phi)
\end{aligned}
\end{equation}

Next, we combine the 3D RGB-T person models and the 3D RGB-T clothing models. In other words, we let the 3D person ``wear" the 3D clothing. The 3D RGB-T person model consists of the 3D body model $ M_{\mathrm{body}}$ , the RGB skin images $P_{\mathrm{skin}}^{\mathrm{vis}}$, and the thermal skin image $P_{\mathrm{skin}}^{\inf}$. Therefore, the rendered visible and thermal images of a person wearing the clothing are given by:

\begin{equation}
\begin{aligned}
I_{\mathrm{person}}^{\mathrm{vis}}=\mathrm{R}(M_{\mathrm{body}},P_{\mathrm{skin}}^{\mathrm{vis}},I_{\mathrm{cloth}}^{\mathrm{vis}}) \\ I_{\mathrm{person}}^{\mathrm{thm}}=\mathrm{R}(M_{\mathrm{body}},P_{\mathrm{skin}}^{\mathrm{thm}},I_{\mathrm{cloth}}^{\mathrm{thm}})
\end{aligned}
\end{equation}

To simulate physical attacks under different environments, we paste the rendered images $I_{\mathrm{person}}^{\mathrm{vis}}$ and $I_{\mathrm{man}}^{\mathrm{thm}}$ onto the aligned visible and thermal background images $I_{\mathrm{back}}^{\mathrm{vis}}$ and $I_{\mathrm{back}}^{\mathrm{thm}}$, respectively. The pasted images are given by:

\begin{equation}
\begin{aligned}
I_{\mathrm{paste}}^{\mathrm{vis}}&=\mathrm{Paste}(I_{\mathrm{person}}^{\mathrm{vis}},I_{\mathrm{back}}^{\mathrm{vis}}) \\
I_{\mathrm{paste}}^{\mathrm{thm}}&=\mathrm{Paste}(I_{\mathrm{person}}^{\mathrm{thm}},I_{\mathrm{back}}^{\mathrm{thm}})
\end{aligned}
\end{equation}

\vspace{-5pt}
\subsection{Loss Functions}
\vspace{-3pt}
We input the pasted images $I_{\mathrm{paste}}^{\mathrm{vis}}$ and $I_{\mathrm{paste}}^{\mathrm{thm}}$ into the RGB-T detection model $f$. The optimization objective of the adversarial pattern is to minimize the RGB-T detector’s confidence score $f_\mathrm{obj}$ of the person wearing the adversarial clothing. Specifically,  the optimization loss function is given by: 
\begin{equation}
\label{equ:loss}
    L=f_\mathrm{obj}(I_{\mathrm{paste}}^{\mathrm{vis}}, I_{\mathrm{paste}}^{\mathrm{thm}}).
\end{equation}

To improve the transferability of the adversarial pattern across different fusion architectures of RGB-T detectors, we propose a fusion-stage ensemble method. 
This approach integrates multiple RGB-T detectors with different fusion architectures, including early-fusion, mid-fusion, late-fusion, and independent RGB-T detectors during optimization.
The ensemble optimization loss function can be formulated as follows:
\begin{equation}
    L_{\mathrm{ensemble}}= w_{1} \cdot L_{\mathrm{early}}+w_{2} \cdot L_{\mathrm{mid}}+w_{3} \cdot L_{\mathrm{late}}+w_{4} \cdot L_{\mathrm{indep}}.
\end{equation}
where $w_{i}, i =1,2,3,4$ are the empirically determined weights for each fusion stage.

\subsection{Physical Implementation} \label{sec:make cloth}

After obtaining the optimized adversarial textures using SDCO method, we print the adversarial textures with RGB color onto fabric. For regions requiring the pasting of aluminum film, the fabric was marked with an ``X'' to indicate their placement. 
Each pixel on the fabric measures 25mm $\times$ 25mm. 
A tailor then processed the printed fabric into clothing, including a shirt and pants, and pasted the aluminum film (only 0.1 mm thick) into the clothing regions marked with ``X''. 
Through this process, we successfully created physical RGB-T adversarial clothing.

\section{Experiments}

\subsection{Dataset}

To simulate RGB-T attacks in diverse real-world environments, we used an aligned RGB-T dataset named FLIR-aligned \cite{zhang2020multispectral}. 
It contains 4,129 well-aligned RGB-T image pairs for training and 1,013 RGB-T image pairs for testing. 
We used this dataset as the source of background images in our 3D RGB-T simulations, and evaluated our method on the first 500 image pairs from the test split.

\subsection{Target RGB-T Detectors} 

\label{sec:target detectors}

We selected typical RGB-T detectors across different fusion architectures as our primary attack targets, including an SOTA early-fusion detector Prob-E \citep{chen2022multimodal}, an SOTA mid-fusion detector Prob-M \citep{chen2022multimodal}, an SOTA late-fusion detector Prob-L \citep{chen2022multimodal}, and SOTA independent RGB-T detectors YOLOv11(RGB) and YOLOv11(T) \citep{khanam2024yolov11}. After performing attacks on these RGB-T detectors in a white-box setting, we transferred our method to unseen black-box RGB-T detectors to evaluate its transferability. These black-box detectors include an early-fusion detector RPN-E \citep{liu2016multispectral}, a mid-fusion detector AR-CNN \citep{Zhang_2019_ICCV}, a late-fusion detector RPN-L \citep{liu2016multispectral}, and independent RGB-T detectors D-DETR(RGB) and D-DETR(T) \citep{zhu2020deformable}. All these detectors achieve high detection confidence scores (above 0.85) on clean RGB-T pedestrian targets, making them strong baselines for evaluating adversarial robustness.

\subsection{Evaluation Metrics}

\begin{table}[bp]
\centering
\small
\caption{Comparison of different methods in the digital world}
\begin{tabular}{l|ccc|cc}
\toprule[1.1pt] 
\multirow{2}{*}{Method} & \multicolumn{5}{c}{ASR (\%) $\uparrow$} \\
\cline{2-6}
& Prob-E & Prob-M & Prob-L & \multicolumn{2}{c}{YOLOv11} \\
& & & & RGB & T \\
\hline
Clean & 0.2 & 0.4 & 0.2 & 0.4 & 0.2 \\
Random & 15.6 & 12.0 & 3.4 & 0.2 & 0.6 \\
MAP\cite{kim2022map} & 31.4 & 37.2 & 11.2 & 6.8 & 4.2 \\
MIC\cite{kim2023multispectral} & 26.2 & 24.0 & 12.4 & 5.8 & 4.0 \\
UAP\cite{wei2023unified} & 25.4 & 27.8 & 5.6 & 2.8 & 4.4 \\
Ours & \textbf{100.0} & \textbf{100.0} & \textbf{99.8} & \textbf{98.8} & \textbf{99.4} \\
\bottomrule[1.1pt] 
\end{tabular}
\label{tab:digital_attack}
\end{table}

In our experiments, we adopted the widely used Attack Success Rate (ASR) as the evaluation metric. ASR is defined as the ratio of the number of undetected target pedestrians to the total number of target pedestrians. Similar to previous works \citep{wei2022adversarial,wei2023unified,ZHU_2021_AAAI,ZHU_2022_CVPR,ZHU_2024_CVPR}, we set the Intersection over Union (IoU) threshold between the predicted bounding box and the ground truth bounding box to 0.5, and the confidence threshold to 0.6. The reported ASR was computed as the average success rate across different viewing angles, distances, and scene conditions.

\subsection{Attack RGB-T Detectors in the Digital World} \label{sec:digital attack}

Based on the method introduced in Sec. \ref{sec:design}, we designed the non-overlapping adversarial RGB-T patterns where each pixel in the pattern measures 10 $\times$ 10 (see \textit{SM} for further analysis of this parameter). Next, we optimized the adversarial patterns using the SDCO method presented in Sec. \ref{sec:DCMA}, with the mask probability $\alpha$ for SRD set to 70\%. A further analysis of $\alpha$ can be found in Sec. \ref{sec:ablation}. Additional experimental settings are detailed in \textit{SM}. After optimization, we obtained the adversarial clothing textures for RGB-T detectors with different fusion architectures, along with the rendered 3D RGB-T human models with adversarial clothing (Fig. \ref{main_process}). See \textit{SM} for the optimized 3D RGB-T models for different RGB-T detectors. 

Next, we applied the rendered 3D RGB-T human model to the test set of the FLIR-aligned dataset, following the process described in Sec. \ref{sec:apply pattern}. 
The generated images were then input into RGB-T detectors with different fusion architectures including Prob-E, Prob-M, Prob-L, and YOLOv11 to evaluate the attack effectiveness of our adversarial pattern in the digital world. 
For a fair comparison, we employed the clean clothing pattern (pure color pattern), random RGB-T pattern (without optimization), and the adversarial patterns generated by previous works including MAP, UAP, and MIC. These patterns were applied according to their original papers and released codes. We used ASR as the evaluation metric. The results are shown in Tab. \ref{tab:digital_attack}. Our method achieved an average ASR of 99.6\% across RGB-T detectors with different fusion stages, while the ASR for the control group was below 37.2\%. A set of visual examples comparing these methods is shown in \textit{SM}.

This indicates that our method effectively attacked RGB-T detectors with different fusion architectures in the digital world, outperforming simple baselines and previous RGB-T attack methods. 

\subsection{Analysis of Adversarial Effects at Different Distances and Angles}

\begin{figure}[bp]
\centering
\includegraphics[width=1\columnwidth]{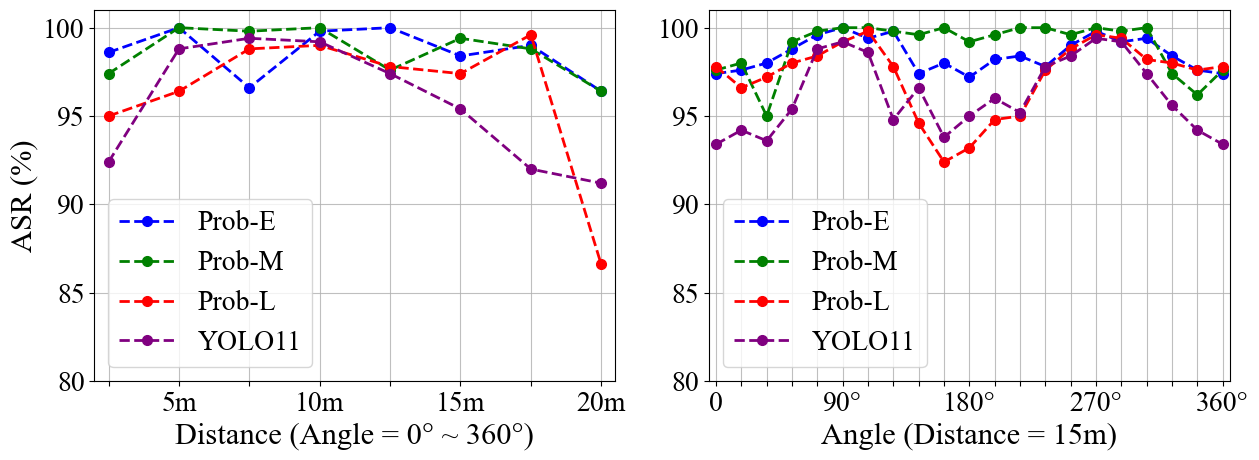} 
\caption{ ASRs for different RGB-T detectors at various (a) distance and (b) viewing angles. }
\label{angle-distance}
\end{figure} 

We further analyzed the ASRs of our method at various viewing angles and distances. The viewing angle varied from 0 to 360 degrees, with samples taken every 18 degrees. The distance ranged from 2.5 meters to 20 meters, with samples taken every 2.5 meters. Fig. \ref{angle-distance} shows the analysis of ASRs for RGB-T detectors with different fusion architectures across various distances and angles.

The results show that our method successfully attacked RGB-T detectors with various fusion architectures at different (full-angle) viewing angles and distances. In comparison, some previous methods \citep{kim2022map,wei2023unified} are only effective within a limited range of angles and distances (e.g., angles from -30 to 30 degrees, distances between 3 to 6 meters). This can be attributed to the fact that our 3D modeling-based approach can simulate attacks over a broader range of angles (full view) and distances, which is a significant improvement over traditional 2D simulation-based methods.

\begin{table}[tbp]
\centering
\small
\caption{Ablation study for SDCO Method}
\begin{tabular}{l|ccc|cc}
\toprule[1.1pt] 
\multirow{2}{*}{Method} & \multicolumn{5}{c}{ASR (\%) $\uparrow$} \\
\cline{2-6}
& Prob-E & Prob-M & Prob-L & \multicolumn{2}{c}{YOLOv11} \\
& & & & RGB & T \\
\hline
w/o SRD & 78.6 & 88.4 & 67.2 & 48.2 & 46.4 \\
w SRD & \textbf{100.0} & \textbf{100.0} & \textbf{99.8} & \textbf{98.8} & \textbf{99.4} \\
\bottomrule[1.1pt] 
\end{tabular}
\label{tab:ablation_DCMA}
\end{table}

\subsection{Ablation Study for SDCO method} \label{sec:ablation}

The core technique of SDCO is Spatially-Random Discretization (SRD). We conducted ablation experiments to evaluate the effectiveness of SRD while keeping all other experimental settings the same as in Sec. \ref{sec:digital attack}. 
The results, shown in Tab. \ref{tab:ablation_DCMA}, indicate that SRD effectively improved the attack effectiveness against RGB-T detectors with different fusion architectures. This is because SRD effectively balances the dual-modality variable optimization.

We further analyzed the impact of the key parameter $\alpha$ of SRD on the ASR, which serves as a balancing factor between the optimization parameters of the visible and thermal modalities. 
The results are detailed in \textit{SM}. 
We observed that the highest average ASR was achieved when $\alpha$ was set to 0.7. Therefore, we set $\alpha = 0.7$ in our experiments.

\subsection{Comparison with Other Optimization Methods}

For simultaneously optimizing discrete and continuous variables, there are other optimization algorithms such as Gumbel-Softmax \citep{conf/iclr/JangGP17} and Straight-Through Estimator \citep{bengio2013estimating} (STE) . Gumbel-Softmax outputs a soft distribution during training and uses a hard distribution during inference, while STE outputs a hard distribution during the forward process and uses a differentiable approximation during the backward process. Both methods transform the discrete optimization problem into an approximate continuous optimization problem, enabling the use of gradient-based methods to optimize both types of variables. We applied these methods to optimize adversarial textures while keeping the settings consistent with Sec. \ref{sec:digital attack}. 

The experimental results, shown in Tab. \ref{tab:compare_DCMA}, indicates that in our experiments, our SDCO method outperforms Gumbel-Softmax and STE. This is likely because Gumbel-Softmax and STE essentially perform continuous optimization and discrete operations in different sequential stage (e.g., training vs. inference, forward vs. backward), while our method simultaneously performs continuous optimization and discrete operations in different spatial aeras, which aligns with the spatial distribution of two types of variables in our RGB-T adversarial clothing.

\begin{table}[tbp]
\centering
\small
\caption{Comparison with other optimization methods}
\begin{tabular}{l|ccc|cc}
\toprule[1.1pt] 
\multirow{2}{*}{Method} & \multicolumn{5}{c}{ASR (\%) $\uparrow$} \\
\cline{2-6}
& Prob-E & Prob-M & Prob-L & \multicolumn{2}{c}{YOLOv11} \\
& & & & RGB & T \\
\hline
Gumbel\cite{conf/iclr/JangGP17} & 78.6 & 87.8 & 60.0 & 34.0 & 22.6 \\
STE\cite{bengio2013estimating} & 95.6 & 96.8 & 94.4 & 92.4 & 86.8 \\
SDCO & \textbf{100.0} & \textbf{100.0} & \textbf{99.8} & \textbf{98.8} & \textbf{99.4} \\
\bottomrule[1.1pt] 
\end{tabular}
\label{tab:compare_DCMA}
\end{table}

\begin{figure*}[htbp]
\centering
\includegraphics[width=1.93\columnwidth]{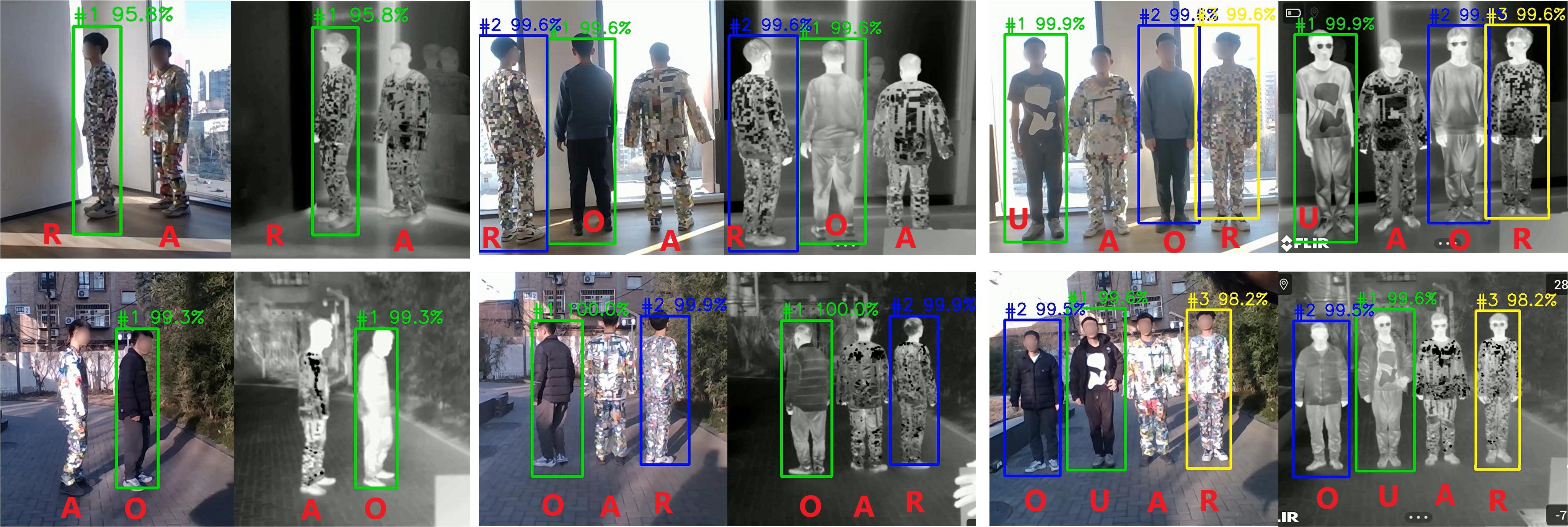} 
\caption{ Visualization of physical RGB-T attacks across diverse scenes. Top row: indoor scenarios. Bottom row: outdoor scenarios. O: ordinary clothes. R: random pattern clothes. U: UAP patch. A: adversarial clothes.  See \textit{SM} for the \textit{Demo Video} and more examples. }
\label{phy_exps}
\end{figure*} 

\subsection{Attack RGB-T Detectors in the Physical World}

\begin{table}[bp]
\centering
\small
\caption{Comparison of different methods in the physical world}
\begin{tabular}{l|ccc|cc}
\toprule[1.1pt] 
\multirow{2}{*}{Method} & \multicolumn{5}{c}{ASR (\%) $\uparrow$} \\
\cline{2-6}
& Prob-E & Prob-M & Prob-L & \multicolumn{2}{c}{YOLOv11} \\
& & & & RGB & T \\
\hline
Clean & 15.2 & 19.6 & 15.3 & 9.4 & 11.6 \\
Random & 15.6 & 21.5 & 15.3 & 8.8 & 9.7 \\
UAP\cite{wei2023unified} & 33.4 & 33.3 & 27.6 & 21.0 & 22.2 \\
Ours & \textbf{73.5} & \textbf{76.5} & \textbf{79.2} & \textbf{61.2} & \textbf{64.4} \\
\bottomrule[1.1pt] 
\end{tabular}
\label{tab:physical_attack}
\end{table}

\begin{table*}[htbp]
\centering
\small
\caption{Transferability in the digital world. The numbers are ASRs.  See \textit{SM} for the physical-world results.}\label{digital_transfer}
\begin{tabular}{c|cccccccc}
\toprule[1.1pt]  
\diagbox{Train}{Test} & Prob-E\cite{chen2022multimodal} & Prob-M\cite{chen2022multimodal} & Prob-L\cite{chen2022multimodal} & YOLOv11\cite{khanam2024yolov11} & RPN-E\cite{liu2016multispectral} & AR-CNN\cite{Zhang_2019_ICCV} & RPN-L\cite{liu2016multispectral} & D-DETR\cite{zhu2020deformable} \\
\hline
Prob-E\cite{chen2022multimodal} & 100.0 & 99.0 & 11.2 & 1.0 & 95.4 & 67.8 & 96.2 & 48.6 \\
Prob-M\cite{chen2022multimodal} & 81.8 & 100.0 & 39.0 & 0.4 & 92.4 & 64.4 & 92.1 & 70.6 \\
Prob-L\cite{chen2022multimodal} & 92.8 & 94.6 & 99.8 & 0.8 & 91.2 & 71.2 & 97.0 & 91.8 \\
YOLOv11\cite{khanam2024yolov11} & 61.0 & 86.4 & 38.2 & 98.4 & 87.0 & 42.0 & 78.6 & 70.4 \\
Ensemble & 99.8 & 100.0 & 99.4 & 96.2 & 94.8 & 76.4 & 97.4 & 99.0 \\
\bottomrule[1.1pt] 
\end{tabular}
\end{table*}

Based on the results of 3D digital simulation, we created physical RGB-T adversarial clothes according to the process described in Sec. \ref{sec:make cloth}. To ensure a fair comparison, we selected ordinary clothing, random RGB-T pattern clothing, latest RGB-T attack method UAP patch as control groups (MAP and MIC were excluded due to the lack of corresponding special materials such as low-E films, which makes physical reproduction difficult). 
See \textit{SM} for the photos of our adversarial clothes and the control groups.

We tested the physical-world attack effectiveness of RGB-T adversarial clothes and the control groups. We invited 5 volunteers to participate in the experiments. The human-related experiments have been approved by the Institutional Review Board (IRB). Volunteers participated in groups of two, three, or four, wearing our RGB-T adversarial clothing, random-pattern RGB-T clothing, ordinary clothing, or holding a UAP patch. We used an iPhone 13 Pro and a FLIR T560 thermal camera to simultaneously capture visible and thermal images of these volunteers. The captured scenes included both indoor and outdoor environments, spanning different times of the day, including morning, noon, afternoon, and nightfall. The camera angles covered a full 0-360$^{\circ}$ range, and the capture distances ranged from 2 to 15 meters.  We recorded 116 videos and sampled 5220 visible-thermal image pairs at a frame rate of 1 frame per second.
The collected images were then input into the corresponding RGB-T detectors for detection, and the ASR was calculated, as shown in Tab. \ref{tab:physical_attack}. 
Some examples are shown in Fig. \ref{one_example} and Fig. \ref{phy_exps}. See \textit{SM} for the \textit{Demo Video}. 

The results indicate that our adversarial clothing successfully evades multiple RGB-T detectors with different fusion architectures across various scenes, angles, and distances, consistently outperforming the baseline methods. 

\subsection{Attack Transferability}

We tested the attack transferability of our RGB-T adversarial clothes to unseen RGB-T detectors, both in the digital and physical worlds. The experimental details are described in \textit{SM}.
It is worth noting that most of the experiments were conducted under a black-box attack setting, which is more challenging but practically significant for real-world applications. The results of the digital experiments are shown in Tab. \ref{digital_transfer}, and the physical world results are available in \textit{SM}. 

The results indicate that that our method successfully attacked various RGB-T detectors in both white-box and black-box settings. More importantly, our fusion-stage ensemble method effectively improved the ASRs against unseen RGB-T detectors compared to patterns optimized for a single model. This suggests that we can just use one single clothing pattern to attack unseen RGB-T detectors with different fusion architectures. 

\subsection{Defense Methods}

We evaluated the effectiveness of eight typical defense methods against our attack method. 
These included five traditional defense techniques: Adversarial Training \citep{goodfellow2014explaining}, Total Variance Minimization \citep{agarwal2021cognitive}, Bit Squeezing \citep{xu2017feature}, JPEG Compression \citep{guo2017countering}, and Pixel Mask \citep{guo2017countering}, along with three state-of-the-art methods specifically designed for defending object detection attacks: PAD \citep{jing2024pad}, NAPGuard \citep{wu2024napguard}, and Jedi \citep{tarchoun2023jedi}. The experimental details of these methods are provided in \textit{SM}. The results show that, although these methods had some defense effects, the ASRs of our method after defense still achieved at least 70\%, further indicating the effectiveness of our attack approach.

\section{Conclusion}

This paper presents a novel method to RGB-T physical attacks using adversarial clothing with NORP.
We construct 3D RGB-T models for human and adversarial clothing to simulate full-view (0$^{\circ}$–360$^{\circ}$) RGB-T attacks. 
NORP is a new adversarial pattern design using distinct visible and thermal materials without overlap, avoiding the light reduction in ORP.
To simultaneously optimizing continuous RGB pixels and discrete thermal pixels within NORP, we propose an SDCO method.
Additionally, we introduce a fusion-stage ensemble method to enhance the transferability of attacks to unseen RGB-T detectors.
Through systematic evaluation in both the digital and physical world,
our work comprehensively reveals the vulnerabilities of RGB-T detectors across different fusion architectures, viewing angles, and distances, which is important for building more robust detectors in the future.

\section{Acknowledgments}
This work was supported by the National Natural Science
Foundation of China (No. U2341228 and No. 62576187).

{
    \small
    \bibliographystyle{ieeenat_fullname}
    \bibliography{main}
}

\end{document}